# Deep-learning based Tools for Automated Protocol Definition of Advanced Diagnostic Imaging Exams

Andrew S. Nencka,  Mohammad Sherafati,  Timothy Goebel,  Parag Tolat and  Kevin M. Koch

*SafeNet Consulting, Milwaukee, WI*
*Department of Radiology, Medical College of Wisconsin, Milwaukee, WI*

## ABSTRACT

**Purpose:** This study evaluates the effectiveness and impact of automated order-based protocol assignment for magnetic resonance imaging (MRI) exams using natural language processing (NLP) and deep learning (DL).

**Methods:** NLP tools were applied to retrospectively process orders from over 116,000 MRI exams with 200 unique sub-specialized protocols ("Local" protocol class). Separate DL models were trained on 70% of the processed data for "Local" protocols as well as 93 American College of Radiology ("ACR") protocols and 48 "General" protocols. The DL Models were assessed in an "auto-protocoling (AP)" inference mode which returns the top recommendation and in a "clinical decision support (CDS)" inference mode which returns up to 10 protocols for radiologist review. The accuracy of each protocol recommendation was computed and analyzed based on the difference between the normalized output score of the corresponding neural net for the top two recommendations.

**Results:** The top predicted protocol in AP mode was correct for 82.8%, 73.8%, and 69.3% of the test cases for "General", "ACR", and "Local" protocol classes, respectively. Higher levels of accuracy over 96% were obtained for all protocol classes in CDS mode. However, at current validation performance levels, the proposed models offer modest, positive, financial impact on large-scale imaging networks.

**Conclusions:** DL-based protocol automation is feasible and can be tuned to route substantial fractions of exams for auto-protocoling, with higher accuracy with more general protocols. Economic analyses of the tested algorithms indicate that improved algorithm performance is required to yield a practical exam auto-protocoling tool for sub-specialized imaging exams.

anencka@mcw.edu (A.S. Nencka)

ORCID(s): Andrew S. Nencka ( 0000-0001-5268-2718); Mohammad Sherafati ( 0000-0001-5362-7550); Timothy Goebel ( ); Parag Tolat ( 0000-0003-1260-5456); Kevin M. Koch ( 0000-0003-4490-9761)

# 1. INTRODUCTION

Health care expenditures in the United States (US) represent more than 19% of its gross domestic product. This is approximately twice that of other high-income countries, with a similar outcome in life expectancy [1, 2, 3]. Advanced imaging, including computed tomography (CT) and magnetic resonance imaging (MRI), strongly contributes to health care costs. Notably, the number of MRI exams per 1,000 residents in the US has doubled over the past two decades[4].

The imbalance between imaging costs and diagnostic benefits is rooted in the inappropriate utilization of imaging resources. Inefficiencies have motivated reforms in imaging utilization which support evidence-based imaging [5] and utilize clinical decision support (CDS) systems [6, 7]. Computerized physician order entry [8, 9] software tools have provided evidence supporting improved process outcomes. These outcomes include both guideline adherence [10] and reduced imaging overuse [11].

Though CDS algorithms offer appropriate "general" imaging protocols (e.g, contrast-enhanced imaging in malignant disease), they do not currently model more complex, advanced sub-specialized imaging protocols that require expert radiologist intervention.

In a value-oriented radiology model, protocol selection, a non-interpretive task, takes place before image acquisition, diagnostic interpretation, and report completion [12]. Protocol selection takes up to 6% of a radiologist's time and is a frequent source of interruptions [13]. It is also labor-intensive, requiring the choice of imaging modality and planes, contrast agent, acquisition parameters, acquired series, and anatomical area covered.

Such complexity causes protocol variation [14] due to individual radiologist practices and preferences. This variation leads to decreased imaging appropriateness, increased interpretive time, and less optimal outcomes. This has motivated the development of tools to standardize protocol construction [15]. Standardization aligns with the principles of Imaging 3.0™: appropriateness, quality, safety, efficiency, and patient experience [16].

Radiologic technologists rarely change protocols from an ordered advanced imaging exam. Furthermore, the overall change rate for radiologists and residents for the most commonly ordered CT and MRI studies is similarly very low. These factors have led experts to conclude that order entry protocol selection is amenable for automation [17].

Machine learning and deep learning (DL) [18] can be leveraged with information from orders and patients' electronic medical records (EMRs)to train artificial intelligence (AI) models to predict protocols from new orders. Robustly auto-protocoling sub-specialized exams could a) reduce the expert effort in protocoling exams, b) enable workflows for patient self-scheduling advanced exams, c) simplify the use of sub-specialized protocols, d) reduce error arising from interrupting radiologist interpretive workflows, and e) expand time for radiologists to perform more valuable tasks.

The growth of AI applications in interpretive and non-iterpretive radiology has been extensively reviewed [19, 20].

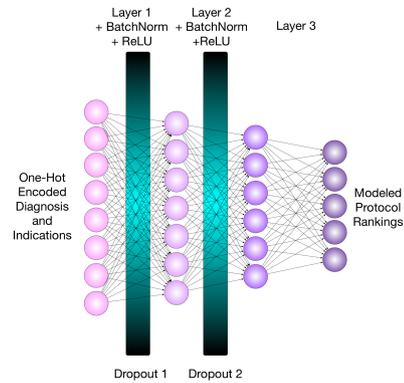

Figure 1: The fully connected neural network utilized in this study for MRI order protocoling. A one-hot encoded input vector of indication and diagnosis words is input, and vector scoring each protocol is output. Dropout and batch normalization layers are included to improve performance with limited training data.

Related to this work, AI-DL-based models have been employed for protocol assignment and quality improvement, demonstrating promising results in neuroradiology [21, 22] and musculoskeletal MRI [23, 24]. Recently, Karla *et al.* [25] demonstrated an AI-augmented workflow to automate CT and MRI protocoling for 69% of orders in a moderately sized sample from a homogeneous clinical environment. They showed >95% concordance with radiologists in those cases, and provide 92% accuracy in top-three protocol suggestion for the remaining cases.

Here, a similar exam protocoling technique is presented and analyzed. This algorithm uses physician-provided indications for imaging and associated *a priori* diagnoses noted in the EMR. The present work holds some key distinctions from the study presented by Karla et al [25]. First, the application in this work is both more specialized and targeted at a diverse patient population. While previous work considered both CT and MRI, the present work exclusively considers MRI examinations from a data bank that is over six times larger than the combined MRI and CT dataset used by Karla *et al.* [25]. Additionally, the considered patient population is extracted from an academic hospital network serving a more diverse urban and suburban population compared to the previously studied sample from a Veterans Administration hospital. Finally, this work provides an economic analysis of the technology to illustrate the impact of implementing such technology in a large imaging network.

# 2. METHODS
## 2.1. Data Source and Preparation

De-identified MRI protocols and matched free-text indications and diagnoses for the 2018 and 2019 calendar years were extracted from the EMRs of an academic hospital network through an IRB-approved data bank. This yielded 116,224 unique imaging exams.

Minimal pre-processing was performed prior to network training to emulate a real-world application. In pre-processing all characters were set to lower case, stop removed using the Natural Language Toolkit (NLTK) "english" corpus module [26]), and punctuation removed.

Due to the preferential use of MRI for neurologic and orthopedic diagnostics, data were class imbalanced across all potential protocols. The train_test_split function in *Scikit-learn* [27] yielded a class-balanced 70%/30% training/test split of the dataset.

For each order, indications and diagnoses were encoded with a "one-hot encoding" scheme [27] of all words identified in those fields of the training set. The encoded vector included the concatenation of separate vectors for the indication field (19,212 elements) and diagnosis field (4,615 elements). Words in the testing set which were not included in the training set were removed.

Table 1: Prevalence of the five most common protocols for each class of protocols in the considered clinical data set.

| General | ACR | Local |
|---|---|---|
| Spine without contrast | Lumbar spine without contrast | Lumbar spine without contrast |
| 24.7 % | 17.6 % | 13.9 % |
| Head with and without contrast | Head without and with contrast | Brain with and without contrast |
| 17.9 % | 11.6 % | 11.6 % |
| Knee without contrast | Cervical spine without contrast | Cervical spine without contrast |
| 8.4 % | 8.6 % | 8.6 % |
| Head without contrast | Knee without contrast | Brain without contrast |
| 8.1 % | 8.4 % | 5.8 % |
| Spine with and without contrast | Head without contrast | Abdomen without and with contrast |
| 7.8 % | 6.8 % | 4.9 % |

### 2.2. Model Design and Training

All 200 "Local" protocols were mapped to 93 protocols defined by the American College of Radiology ("ACR") "Appropriateness Criteria" ® [28]. Subsequently, each ACR protocol was mapped to one of 48 more "General" protocols. A full mapping of protocols is provided in the supplemental material.

Three fully connected neural networks were designed to ingest the indication and diagnosis vector and return a vector of $n$ scores, where $n$ is the number of protocols in the protocol class. The recommended protocol is the element with the greatest returned value. Figure 1 illustrates this neural network. It included 23,827 inputs (corresponding to each encoded word), and three dense layers with $4n$, $2n$, and $n$ neurons [29]. The network also utilized rectified linear unit (ReLU) [30] activations, 50% training dropout [31] layers, and batch normalization [32].

One network was trained for each set of protocols using an NVIDIA (Santa Clara, California) Titan V graphical processing unit (GPU), an ADAM optimizer with a learning rate of 0.0001 [33], a cross-entropy loss function [34], and two hundred training epochs with data shuffling. Using five data loading threads and a batch size of 24 orders, each training epoch lasted approximately 32 seconds. Inference on test data with a batch size of 24 orders took 0.7 ms per order.

### 2.3. Modes of Operation and Performance Metrics

Two modes of inference use were defined: auto-protocoling (AP) and clinical decision support (CDS). As an AP tool, the top recommendation was returned as the selected protocol. In the CDS mode, the top five recommendations were returned as a set of proposed protocols to be evaluated by a protocoling radiologist/technologist.

Relative amplitude differences of inferred protocol weightings were used to switch between AP and CDS modes. Each output vector was scaled to have a minimum of zero and a total sum of 1.0. The difference between the top recommendation and the second recommendation was computed, and is defined as "delta" which is shown in Fig. 2. The delta can be interpreted as a rough indicator of confidence in the recommended protocol, with a larger value of the delta corresponding to greater confidence for the selected protocol. All reported metrics were computed as a function of delta values.

A rudimentary analysis of the economic impact of using these models was performed based upon the percentage of exams that are routed through the AP model. This analysis assumed an average hourly rate of $38 (technologist) and $206 (radiologist) (salary.com), an average time to protocol an exam of two minutes (technologist) and one minute (radiologist), and an annual volume of 58,000 exams. Cost savings were computed as the fraction of auto-protocoled exams multiplied by the annual exam volume, time to protocol an exam, and practitioner's hourly rate.

## 3. RESULTS

Each model was used to inference the test data set of 34,868 orders. Each protocol was assigned a percentile score for the order. The known protocol acquired in the test data set was then compared to the percentile rank for that protocol in the inferred protocol vector. Histograms of percentile rankings of correct protocols for each of the three AP models are illustrated in Fig. 3.

The delta threshold can be tuned depending upon a site's preference for acceptable levels of discordance between AP and manual results along with the volume of orders routed through the AP workflow. For a given magnitude of the delta, in Fig. 4, all the "AP−" exams below the threshold were protocoled incorrectly and would be appropriately routed to CDS mode for radiologist protocoling. Similarly, "AP+" orders above this threshold were auto-protocoled correctly and would utilize the AP workflow, as desired.

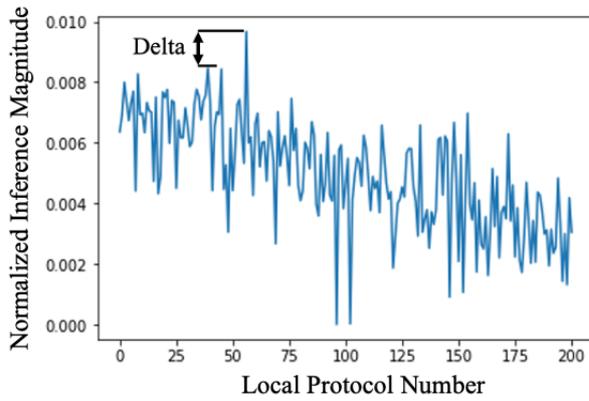

Figure 2: Normalized output of the "Local" model. The magnitude of the rough confidence metric, "delta," is shown as the difference between the top two inferred magnitudes.

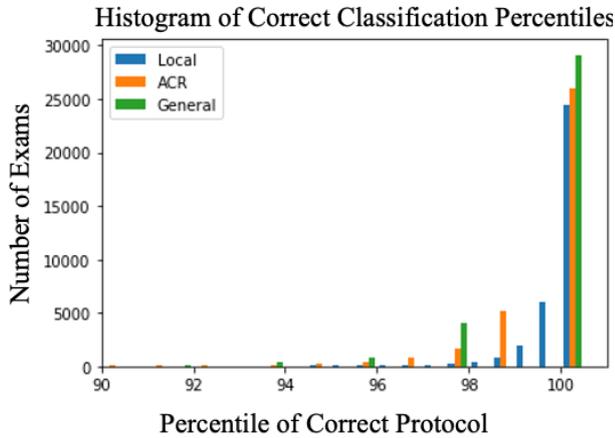

Figure 3: Histograms of percentile rankings of correct protocols for each AP model.

Table 2: Accuracy of the modeled recommended protocols based on a test dataset of 34,868 orders. If used in pure auto-protocoling mode, the top recommendation indicates the percentage of exams wherein the auto-protocoling results were correct. If used in pure clinical decision support (CDS) mode, the top five recommendation indicates the percentage of exams wherein the desired protocol is in the list of returned protocols. If CDS were expanded to return 10 protocols, the final row illustrates the resulting accuracy.

| Accuracy Metric \ Protocol Class | General | ACR | Local |
|---|---|---|---|
| Top Recommendation | 82.8% | 73.8% | 69.3% |
| Top Five Recommendation | 98.6% | 97.0% | 96.0% |
| Top 10 Recommendation | 99.4% | 98.7% | 98.0% |

A well-performing AP model yields a bimodal histogram with larger areas of AP+ and AP− distributions lying above and below the threshold, respectively. For example, histograms shown in Fig. 4 with a shift of the "AP+" distribution towards higher deltas with more general protocols suggest that generalized protocols are better suited for the AP workflow.

Figure 5 illustrates the increasing accuracy of AP models as a function of the delta. With small delta values, the performance of the AP mode is improved with more general sets of protocols. This trend is reversed with higher delta values. If a high rate of accuracy is desired for AP workflows, more protocols that are more specific and are thresholded with higher delta values are preferable, although this results in a low percentage of exams that proceed through the AP workflow.

The percent of cases with the correct protocol in the top five recommendations of the models' CDS mode is illustrated in Fig. 6. With a smaller set of general protocols, there is a higher probability of the correct protocol being in the top recommendations, and the "General" model outperforms the other models across all considered delta thresholds. As delta increases, the performance of the CDS mode improves. The highly specialized "Local" model outperforms the more general "ACR" protocoling model with a delta threshold between 0.0006 and 0.0020.

### 3.1. Economic Impact Analysis

The fraction of exams that are routed to AP mode for each protocol class monotonically decreases with the increase of delta as shown in Fig. 7. This decay is faster for the sub-specialized protocols. For general protocols, a higher volume of exams can be routed through the auto-protocoling workflow irrespective of the delta. For sufficiently small delta ($\sim 0.0002$), over 80% of exams can be routed to AP mode, irrespective of the model utilized.

Economic impact analysis of these results suggests that modest cost savings can be achieved using these DL-based protocoling algorithms. Approximately 25% of a technologist or 12.5% of a radiologist full time equivalent employee (FTE) can be saved with a delta threshold of 0.10 using the sub-specialized "Local" model which achieves over 95% accuracy in its CDS mode and nearly 95% accuracy in its auto-protocoling mode. While these savings are not inconsequential, they are not substantially impactful on the budget of the large healthcare imaging network utilized for the present study.

More substantial savings could be achieved if less specialized protocols are used, enabling the auto-protocoling workflow to be employed more often, with the "General" model achieving over 90% accuracy in AP mode, 97% accuracy in CDS mode, and over 150% or 37% technologist or radiologist FTE savings with a delta threshold of 0.0016.

Overall, these results highlight the need for future work to render the evaluated technologies viable in clinical enterprises relying on specialized exam protocols.

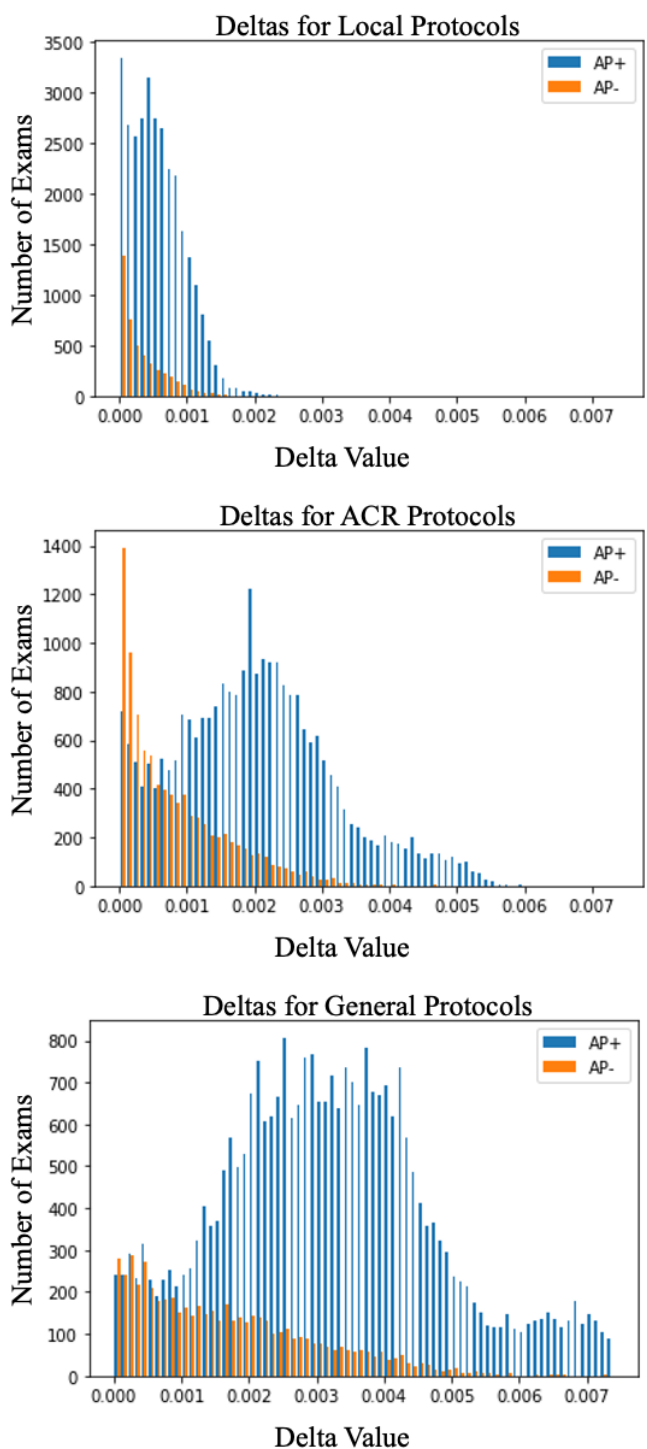

Figure 4: Histograms of delta for each AP model when the AP recommendation was correct (AP+) and incorrect (AP-)

## 4. DISCUSSION

In this feasibility study, AI-based algorithms were developed and tested to automate order-based protocoling for 116,224 MRI exams ordered during 2018 and 2019 in an urban and suburban imaging network.

Table 2 illustrates that the evaluated models are not sufficient as pure AP tools. This is likely due to the diverse patient population and limited information contained in physician order notes. If used for protocol recommendation in a CDS capacity, wherein five to 10 protocols are recommended to an individual protocoling the exam, the models perform very well. Furthermore, if generalized exam protocols are utilized, model performance improves. The results shown in Figs. 3 through 7 suggest the use of an alternative hybrid solution, wherein AP is performed if a key metric is above threshold and CDS is offered below this threshold.

Rudimentary economic impact analysis of the singular AP/CDS modes evaluated in this study predicts modest cost savings when implementing such approaches. This economic impact performance is highly dependent on the algorithmic performance in the AP mode on the specialized exam protocol pool.

The accuracy of the AP workflow for specialized protocols shown in this work is lower than in other work and the observed delta threshold is not as sharp as that reported in Kalra et al[25], which focused on a population of primarily older males in a single hospital in the Veterans Administration system. In contrast, the present study utilized data from a large urban-based academic medical system serving a broad range of patients. Of note, the broader and more diverse dataset utilized in this study, along with the categorical analysis of exam protocols with increasing specificity, enabled the practical economic analysis that is unique to the present study.

The primary errors in the AP mode of the evaluated models in this study occurred when predicting highly specific protocols. In protocols that were lateralized (i.e. right versus left wrist), protocol errors primarily indicated the appropriate anatomy and contrast agent usage while incorrectly indicating laterality. This could be addressed by adding an ordering question to the electronic medical record software to indicate exam laterality. In protocols which were not lateralized, and included options regarding contrast agent inclusion (i.e. without, with and without, or with contrast), most errors again included correct anatomy and incorrect contrast agent selection. There are cases wherein the inclusion of contrast agent would be optimal for imaging but detrimental for the patient (i.e. pregnant patients and patients with acute renal injury [35]), which may contribute to this error. The addition of model inputs based upon the medical record to gain information regarding contraindications for contrast agent usage could further improve model performance.

The financial analysis included in this study highlights the need for additional informatics integration, such as the aforementioned order laterality and contrast agent specifications, that must be utilized in combination with the demonstrated DL-based algorithms to improve performance. Without improved inputs to AP algorithms, it is unlikely that such tools will offer a substantial benefit to radiology practices. However, it is noted that the addition of such information to the neural network inputs is not a substantial roadblock and is likely to provide a viable path forward in the development of AP workflows.

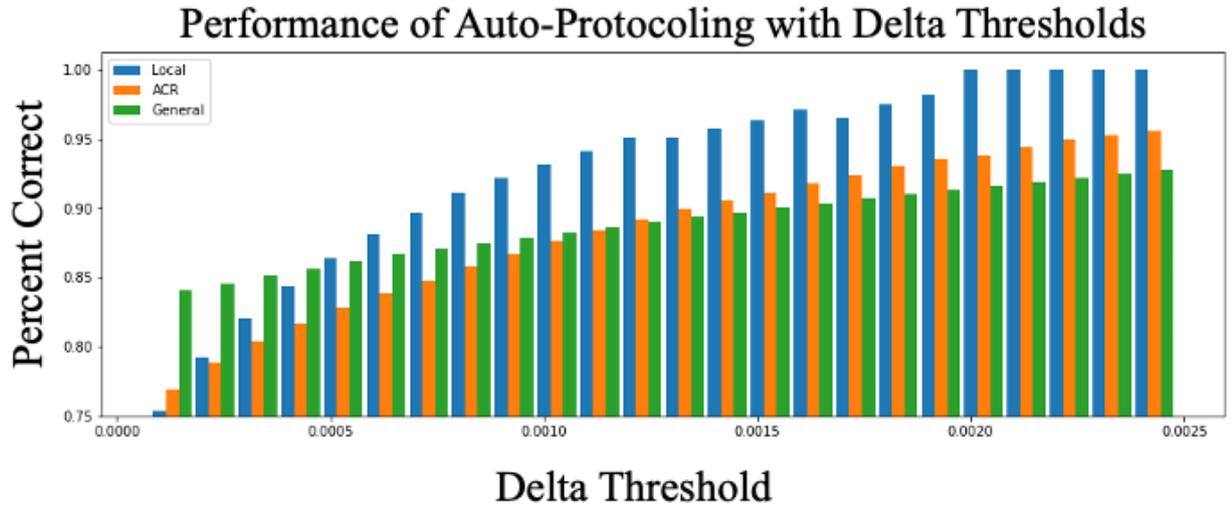

Figure 5: Bar chart of percentage of correct AP recommendations as a function of delta for each model

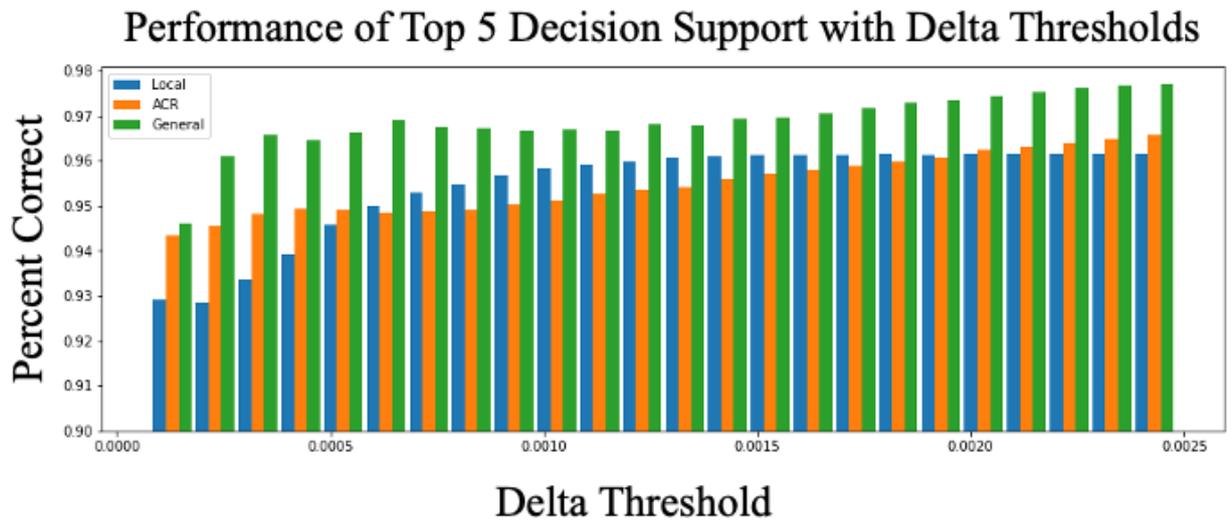

Figure 6: Bar chart of percentage of CDS recommendations containing the correct protocol as a function of delta for each model

The present study shows that DL-based AP technology can be designed to relieve a portion of the manual protocoling burden which can interrupt interpretive work. Such interruptions cause frustrations and lead to increased exam reporting times or potential errors in image interpretations [36, 37, 38, 13, 39]. In addition, such a tool may provide radiologists with increased uninterrupted time. Recent work supports this hypothesis[40], whereby a simple AP intervention providing a standardized CT protocoling system for emergency exams can improve the radiology workflow by reduc-ing the lag time between ordering and protocoling, and improving time to the final report. Moreover, that work demonstrated improvements in radiology residents' work satisfaction and wellness when such tools are utilized.

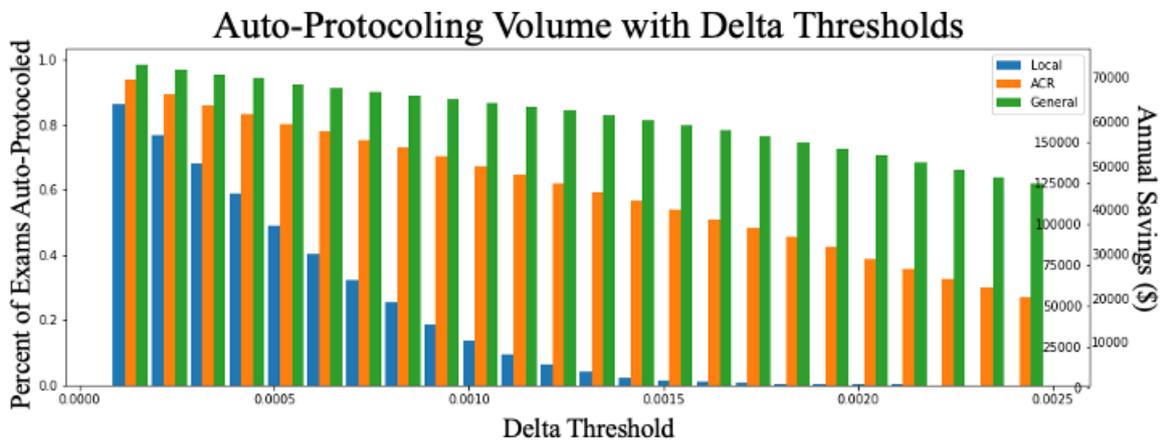

Figure 7: Bar chart of the AP model case volume. The vertical axis on the left indicates the fraction of exams in the test data set which are routed through the AP model for the "delta" threshold indicated along the horizontal axis. The right vertical axis indicates the estimated annual cost reduction associated with the auto-protocoling tool usage if radiologists primarily perform protocoling duties (scale to left of axis) or technologists (scale to right of axis).

The present study has two main limitations: 1) source data was limited to a single hospital network and 2) neural network evaluation was performed on retrospective data. While the trained models are not necessarily appropriate for application to data in a different hospital network, the model architecture, bifurcated workflow, and delta analysis are directly transferable. As a result, real-time assessment of discrepancies between radiologist prescriptions and algorithm recommendations is not captured, and temporal savings associated with a CDS list of proposed protocols is not measured. Neural network hyperparameters were determined through a cursory grid search, and more formal consideration of network architecture could hold value. Further accuracy improvements are feasible through the increases in data extracted from the EMR for model inputs. The economic analysis performed in this study only accounts for the time spent protocoling of exams. It does not take into account further benefits of the evaluated technology, such as improved interpretive productivity and reduced protocoling time with CDS usage.

## 5. CONCLUSIONS

Exam protocol selection for advanced diagnostic imaging exams is a monotonous and unbilled use of professional resources. The present study demonstrates that DL-based AP and CDS tools offer the potential to alleviate these resource demands. By tuning a threshold for switching between AP and CDS modes, compromises between cost savings and protocol accuracy can be modulated. Despite the encouraging preliminary demonstration of DL-based AP of advanced imaging exams, economic analyses suggest that substantially improved algorithm performance will be required to yield a practical AP tool for sub-specialized imaging exams. Specific additional inputs to the models, including laterality and patient-specific contrast contraindications, are hypothesized to enable such performance improvements.

**TAKE-HOME POINTS**

- Deep learning-based automated diagnostic imaging exam protocoling is feasible in a diverse clinical network.

- The performance of such algorithms depends on the level of specificity of pre-defined exam protocols.

- The output of AP algorithms can be thresholded to modulate exam volume between AP and CDS tools.

- Economic analysis of purely order-driven models suggests that increased levels of information will be required as inputs to trained models before such algorithms are viable for clinical usage.

## 6. ACKNOWLEDGMENTS

The authors thank Robin Ausman and Bradly Swearingen for their work to extract the data utilized in this analysis, as well as Jess Zieman and Jeff Rehm for support regarding the financial aspects of this work.

## 7. BACKMATTER

| General Protocol Title | ACR Protocol Title | Local Protocol Title |
|---|---|---|
| head w/ w/o | mri head without and with iv contrast | mr brain wo + w cont |
| head w/ w/o | mri orbits without and with iv contrast | mr orbit(s) w/o + w cont |
| head w/ w/o | mri sella without and with iv contrast | mr sella w/o + w cont |
| head w/ w/o | mri head without and with iv contrast | mr brain stealth wo + w cont fh |
| head w/ w/o | mri head without and with iv contrast | mr sim brain with interp |
| head w/ w/o | mri head without and with iv contrast | mr brain gamma wo + w cont fh |
| head w/ w/o | mri head without and with iv contrast | mr brain intraop w cont w rad interp fh |
| head w/ w/o | mri head without and with iv contrast | ct preop head stealth mrk w cont |
| head w/ w/o | mri head perfusion with iv contrast | mr rcbv sequence |
| head w/ w/o | mri head and internal auditory canal without and with iv contrast | mr iac w/o + w cont |
| head w/ w/o | mrv head without and with iv contrast | mr mrv brain w/o & with cont |
| head w/ w/o | mra head without and with iv contrast | mr mra head w/o + w cont |
| head w/ w/o | mri orbit face neck without and with iv contrast | mr face w/o + w cont |
| head w/ w/o | mri orbit face neck without and with iv contrast | mr sim facial with interp |
| head w/ w/o | mri head without iv contrast with dti | mr brain w/o+w cont + dti |
| knee w/o | mri knee without iv contrast | mr knee lt w/o cont |
| knee w/o | mri knee without iv contrast | mr knee rt w/o cont |
| knee w/o | mri knee without iv contrast | mr pre-surg knee planning rt |
| knee w/o | mri knee without iv contrast | mr pre-surg knee planning lt |
| spine w/o | mri lumbar spine without iv contrast | mr l spine w/o cont |
| spine w/o | mri lumbar spine without iv contrast | mr l spine w/o cont chiro read cdi |
| spine w/o | mri thoracic and lumbar spine without iv contrast | mr t spine w/o cont |
| spine w/o | mri cervical spine without iv contrast | mr c spine w/o cont |
| spine w/o | mri cervical spine without iv contrast | mr c spine w/o cont chiro read cdi |
| spine w/ w/o | mri lumbar spine without iv contrast | mr l spine w/o + w cont |
| spine w/ w/o | mri thoracic and lumbar spine without and with iv contrast | mr t spine w/o + w cont |
| spine w/ w/o | mri thoracic and lumbar spine without and with iv contrast | mr sim thoracic spine with interp |
| spine w/ w/o | mri cervical spine without and with iv contrast | mr c spine w/o + w cont |
| spine w/ w/o | mri lumbar spine without and with iv contrast | mr sim lumbar spine with interp |
| shoulder w/o | mri shoulder without iv contrast | mr shoulder lt w/o cont |
| shoulder w/o | mri shoulder without iv contrast | mr shoulder rt w/o cont |
| shoulder w/o | mri shoulder without iv contrast | mr scapula lt w/o cont |
| shoulder w/o | mri shoulder without iv contrast | mr scapula rt w/o cont |
| shoulder w/o | mri brachial plexus without iv contrast | mr brach plex rt w/o cont |
| shoulder w/o | mri brachial plexus without iv contrast | mr brach plex lt w/o cont |
| hand w/o | mri hand without iv contrast | mr hand lt w/o cont |
| hand w/o | mri hand without iv contrast | mr finger lt w/o cont |
| hand w/o | mri hand without iv contrast | mr finger rt w/o cont |
| hand w/o | mri hand without iv contrast | mr hand rt w/o cont |
| cardiac w/ w/o | mri heart function and morphology without and with iv contrast | mr cardiac morph-func w/o + w cont |
| cardiac w/ w/o | mri heart with stress without and with iv contrast | mr cardiac morph func w/stress w/o + w cont fh |
| head w/o | mri head without iv contrast | mr brain w/o cont |
| head w/o | mri head without iv contrast | mr memory loss brain w/o cont w 3d imaging |
| head w/o | mri head without iv contrast | mr neuroreader brain w/o cont w 3d imaging |
| head w/o | mri head without iv contrast | mr tmj bilat |
| head w/o | mri head without iv contrast | mr brain stealth wo cont fh |
| head w/o | mri head without iv contrast | ct preop head stealth mrk w/o cont |
| head w/o | mri head without iv contrast | mr brain w/o cont ltd research fh |
| head w/o | mri head without iv contrast | mr brain intraop w/o cont with rad interp fh |
| head w/o | mri head without iv contrast | mr tmj lt |
| head w/o | mri head without iv contrast | mr meg brain w/o cont |
| head w/o | mri head without iv contrast | mr tmj rt |
| head w/o | mri head without iv contrast | ct head stealth mrk w/o cont |
| head w/o | mri head without iv contrast | mr brain gamma wo cont fh |
| head w/o | mra head without iv contrast | mr mra head w/o cont |
| head w/o | mri orbits without iv contrast | mr orbit(s) w/o cont |
| head w/o | mri functional (fmri) head without iv contrast | mr functional tech testing fh |
| head w/o | mr spectroscopy head without iv contrast | mr spectroscopy |
| head w/o | mrv head without iv contrast | mr mrv brain w/o cont. |
| head w/o | mri orbit face neck without iv contrast | mr face w/o cont |
| foot w/ w/o | mri foot without and with iv contrast | mr foot lt w/o + w cont |
| foot w/ w/o | mri foot without and with iv contrast | mr foot rt w/o + w cont |
| foot w/ w/o | mri foot without and with iv contrast | mr toes lt w/o + w cont |
| foot w/ w/o | mri foot without and with iv contrast | mr forefoot rt w/o + w cont |
| foot w/ w/o | mri foot without and with iv contrast | mr forefoot lt w/o + w cont |
| foot w/ w/o | mri foot without and with iv contrast | mr toes rt w/o + w cont |
| foot w/ w/o | mri ankle and hindfoot without and with iv contrast | mr heel rt w/o + w cont |
| foot w/o | mri foot without iv contrast | mr foot rt w/o cont |
| foot w/o | mri foot without iv contrast | mr foot lt w/o cont |
| foot w/o | mri foot without iv contrast | mr forefoot lt w/o cont |
| foot w/o | mri foot without iv contrast | mr forefoot rt w/o cont |
| foot w/o | mri foot without iv contrast | mr toes rt w/o cont |
| foot w/o | mri foot without iv contrast | mr toes lt w/o cont |
| foot w/o | mri ankle and hindfoot without iv contrast | mr hindfoot lt w/o cont |
| foot w/o | mri ankle and hindfoot without iv contrast | mr hindfoot rt w/o cont |
| foot w/o | mri ankle and hindfoot without iv contrast | mr heel rt w/o cont |
| foot w/o | mri ankle and hindfoot without iv contrast | mr heel lt w/o cont |
| shoulder w/ | mr arthrography shoulder | mr shoulder lt w cont |
| shoulder w/ | mr arthrography shoulder | mr shoulder rt w cont |
| pelvis w/ w/o | mri sacroiliac joints without and with iv contrast | mr sacrum w/o + w contrast |
| pelvis w/ w/o | mri pelvis without and with iv contrast | mr prostate w/o + w cont |
| pelvis w/ w/o | mri pelvis without and with iv contrast | mr pelvis msk w/o + w cont |
| pelvis w/ w/o | mri pelvis without and with iv contrast | mr pelvis organ w/o + w cont |
| pelvis w/ w/o | mri pelvis without and with iv contrast | mr sim pelvis with interp |
| pelvis w/ w/o | mr enterography | mr enterography w/o + w cont |
| pelvis w/ w/o | mri hip without and with iv contrast | mr hip rt w/o + w cont |
| pelvis w/ w/o | mri hip without and with iv contrast | mr hip lt w/o + w cont |
| pelvis w/ w/o | mra abdomen and pelvis with iv contrast | mr mra pelvis |
| pelvis w/o | mri hip without iv contrast | mr hip lt w/o cont |
| pelvis w/o | mri hip without iv contrast | mr hip rt w/o cont |
| pelvis w/o | mri sacroiliac joints without iv contrast | mr sacrum w/o |
| pelvis w/o | mri sacroiliac joints without iv contrast | mr sacrum w/o chiro read cdi |
| pelvis w/o | mri pelvis without iv contrast | mr pelvis msk w/o cont |
| pelvis w/o | mri pelvis without and with iv contrast | mr pelvis ltd fibroid protocol cdi |
| pelvis w/o | mri pelvis without iv contrast | mr pelvis organ w/o cont |
| abdomen w/ w/o | mri abdomen without and with iv contrast | mr abd w/o + w cont |
| abdomen w/ w/o | mri abdomen without and with iv contrast | mr sim abd with interp |
| abdomen w/ w/o | mri abdomen and pelvis without and with iv contrast | mr abd/pelvis w + w/o cont |
| abdomen w/ w/o | mra abdomen without and with iv contrast | mr mra abdomen w/o + w cont |
| abdomen w/ w/o | mr elastography abdomen | mr abd w + w/o + elastography |
| thigh w/o | mri lower extremity area of interest (not pelvis or hip) without iv contrast | mr femur lt w/o cont |
| thigh w/o | mri lower extremity area of interest (not pelvis or hip) without iv contrast | mr femur rt w/o cont |
| extremity w/o | mri lower extremity area of interest (not pelvis or hip) without iv contrast | mr tib/fib rt w/o cont |
| extremity w/o | mri lower extremity area of interest (not pelvis or hip) without iv contrast | mr low ext lt multi w/o cont |
| extremity w/o | mra lower extremity without iv contrast | mr tib/fib lt w/o cont |
| extremity w/o | mri extremity area of interest without iv contrast | mr humerus rt w/o cont |
| extremity w/o | mri extremity area of interest without iv contrast | mr humerus lt w/o cont |
| extremity w/o | mri extremity area of interest without iv contrast | mr forearm lt w/o cont |
| extremity w/o | mri extremity area of interest without iv contrast | mr forearm rt w/o cont |
| extremity w/o | mra extremity area of interest without and with iv contrast | mr sim ext upper with interp |
| neck w/ | mra neck with iv contrast | mr mra neck w cont |
| breast w/ w/o | mri breast without and with iv contrast bilateral | bi mr breast bilat w/o+w cont with cad |
| breast w/ w/o | image-guided core biopsy breast | bi mr guide breast bx w w/o clip+spec 1st lesion rt |
| breast w/ w/o | image-guided core biopsy breast | bi mr guide breast bx w w/o clip+spec 1st lesion lt |
| breast w/ w/o | image-guided core biopsy breast | bi mr guide breast bx w w/o clip+spec 1st lesion bi |
| breast w/ w/o | mri breast without and with iv contrast | bi mr breast lt w/o+w cont with cad |
| breast w/ w/o | mri breast without and with iv contrast | bi mr breast rt w/o+w cont with cad |
| ankle w/o | mri ankle without iv contrast | mr ankle lt w/o cont |
| ankle w/o | mri ankle without iv contrast | mr ankle rt w/o cont |
| ankle w/o | mri ankle and hindfoot without iv contrast | mr ankle achilles lt w/o cont |
| ankle w/o | mri ankle and hindfoot without iv contrast | mr ankle achilles rt w/o cont |
| wrist w/o | mri wrist without iv contrast | mr wrist rt w/o cont |
| wrist w/o | mri wrist without iv contrast | mr wrist lt w/o cont |
| pelvis w/ | mr arthrography hip | mr hip rt w cont |
| pelvis w/ | mr arthrography hip | mr hip lt w cont |
| pelvis w/ | mri pelvis without and with iv contrast | mr pelvis organ w/cont |
| pelvis w/ | mra extremity area of interest without and with iv contrast | mr mra pelvis + low ext bilat run off w cont |
| cardiac w/o | mri heart function and morphology without iv contrast | mr cardiac morph + func w/o |
| shoulder w/ w/o | mri brachial plexus without and with iv contrast | mr brach plex lt w/o + w cont |
| shoulder w/ w/o | mri brachial plexus without and with iv contrast | mr brach plex rt w/o + w cont |
| shoulder w/ w/o | mri shoulder without and with iv contrast | mr scapula lt w/o + w cont |
| shoulder w/ w/o | mri shoulder without and with iv contrast | mr shoulder rt w/o + w cont |
| shoulder w/ w/o | mri shoulder without and with iv contrast | mr shoulder lt w/o + w cont |
| shoulder w/ w/o | mri shoulder without and with iv contrast | mr scapula rt w/o & with cont |
| extremity w/ w/o | mri extremity area of interest without and with iv contrast | mr humerus rt w/o + w cont |
| extremity w/ w/o | mri extremity area of interest without and with iv contrast | mr humerus lt w/o + w cont |
| extremity w/ w/o | mri extremity area of interest without and with iv contrast | mr forearm rt w/o & with cont |
| extremity w/ w/o | mri extremity area of interest without and with iv contrast | mr forearm lt w/o + w cont |
| extremity w/ w/o | mri extremity area of interest without and with iv contrast | mr mra upp ext rt multi w/o + w cont |
| extremity w/ w/o | mri lower extremity area of interest (not pelvis or hip) without and with iv contrast | mr tib/fib lt w/o + w cont |
| extremity w/ w/o | mri lower extremity area of interest (not pelvis or hip) without and with iv contrast | mr tib/fib rt w/o + w cont |
| extremity w/ w/o | mri lower extremity area of interest (not pelvis or hip) without and with iv contrast | mr low ext lt multi w/o + w cont |
| extremity w/ w/o | mri lower extremity area of interest (not pelvis or hip) without and with iv contrast | mr sim ext lower with interp |
| extremity w/ w/o | mra lower extremity without and with iv contrast | mr mra low ext lt multi w/o + w cont |
| neck w/ w/o | mri neck without and with iv contrast | mr neck soft tissue w/o + w cont |
| neck w/ w/o | mri neck without and with iv contrast | mr sim neck with interp |
| spine w/ | mri lumbar spine without and with iv contrast | mr l spine w cont |
| spine w/ | mri cervical spine with iv contrast | mr c spine w cont |
| spine w/ | mri thoracic and lumbar spine without and with iv contrast | mr t spine w cont |
| elbow w/o | mri elbow without iv contrast | mr elbow rt w/o cont |
| elbow w/o | mri elbow without iv contrast | mr elbow lt w/o cont |
| whole body w/o | mri whole body without iv contrast | mr whole body survey |
| thigh w/ w/o | mri lower extremity area of interest (not pelvis or hip) without and with iv contrast | mr femur rt w/o + w cont |
| thigh w/ w/o | mri lower extremity area of interest (not pelvis or hip) without and with iv contrast | mr femur lt w/o + w cont |
| chest w/o | mri chest without iv contrast | mr chest w/o cont |
| chest w/o | mra chest without iv contrast | mr mra chest w/o cont |
| mrcp | mri abdomen without and with iv contrast with mrcp | mr mrcp w/o + w cont |
| mrcp | mri abdomen without iv contrast with mrcp | mr mrcp w/o cont |
| abdomen w/o | mri abdomen without iv contrast | mr abd w/o cont |
| abdomen w/o | mri abdomen without iv contrast | mr abd w/o cont ltd for iron content fh |
| abdomen w/o | mri abdomen without iv contrast | mr abd w/o cont ltd research fh |
| head w/ | mrv head with iv contrast | mr mrv brain with cont |
| head w/ | mri orbit face neck without and with iv contrast | mr orbit(s) w cont |
| head w/ | mri head with iv contrast | mr brain w cont |
| head w/ | mri head without and with iv contrast | ct head stealth mrk w cont |
| head w/ | mri head without and with iv contrast | mr brain gamma w cont fh |
| neck w/o | mra neck without iv contrast | mr mra neck w/o cont |
| neck w/o | mri neck without iv contrast | mr neck w/o cont |
| ankle w/ w/o | mri ankle without and with iv contrast | mr ankle rt w/o + with cont |
| ankle w/ w/o | mri ankle without and with iv contrast | mr ankle lt w/o + with cont |
| wrist w/ | mr arthrography wrist | mr wrist rt w cont |
| wrist w/ | mr arthrography wrist | mr wrist lt w cont |
| chest w/ w/o | mri chest without and with iv contrast | mr chest w/o + w cont |
| chest w/ w/o | mri chest without and with iv contrast | mr sim chest with interp |
| chest w/ w/o | mra chest without and with iv contrast | mr mra chest w/o + w cont |
| hand w/ w/o | mri hand without and with iv contrast | mr hand rt w/o + w cont |
| hand w/ w/o | mri hand without and with iv contrast | mr hand lt w/o + w cont |
| hand w/ w/o | mri hand without and with iv contrast | mr finger lt w/o + w cont |
| hand w/ w/o | mri hand without and with iv contrast | mr finger rt w/o + w cont |
| knee w/ w/o | mri knee without and with iv contrast | mr knee lt w/o + w cont |
| knee w/ w/o | mri knee without and with iv contrast | mr knee rt w/o + w cont |
| knee w/ w/o | mra knee without and with iv contrast | mr mra bilat knee popliteal entrapment wo + w cont |
| wrist w/ w/o | mri wrist without and with iv contrast | mr wrist lt w/o + w cont |
| wrist w/ w/o | mri wrist without and with iv contrast | mr wrist rt w/o + w cont |
| breast w/o | mri breast without iv contrast bilateral | bi mr breast bilat w/o cont |
| breast w/o | image-guided transthoracic needle biopsy | mr guide needle placement |
| breast w/o | mri breast without iv contrast | bi mr breast rt w/o cont |
| elbow w/ | mr arthrography elbow | mr elbow lt w cont |
| elbow w/ | mr arthrography elbow | mr elbow rt w cont |
| elbow w/ w/o | mri elbow without and with iv contrast | mr elbow rt w/o + w cont |
| elbow w/ w/o | mri elbow without and with iv contrast | mr elbow lt w/o + w cont |
| abdomen w/ | mra abdomen without and with iv contrast | mr mra abdomen w cont |
| knee w/ | mr arthrography knee | mr knee rt w cont |
| knee w/ | mr arthrography knee | mr knee lt w cont |
| chest w/ | mra chest without and with iv contrast | mr mra chest w cont |
| ankle w/ | mr arthrography ankle | mr ankle lt w cont |
| hand w/ | mr arthrography wrist | mr hand rt w cont |
| hand w/ | mr arthrography wrist | mr hand lt w cont |

Table 3: Local protocol titles with corresponding ACR and general protocol titles.